\documentclass[10pt, twocolumn, twoside, a4paper]{IEEEtran}
\usepackage{cite}
\usepackage[pdftex]{graphicx}
\usepackage[cmex10]{amsmath}
\usepackage{amsfonts}
\usepackage{amssymb}
\usepackage{amsbsy}
\usepackage{amsthm}
\usepackage[mathscr]{eucal}
\usepackage{subfigure}

\usepackage{comment}
\usepackage{paralist}
\usepackage{boxedminipage}
\usepackage{eucal}
\usepackage{xcolor}
\usepackage{multirow}

\usepackage{latexsym}
\usepackage{amsmath}
\usepackage[dvips]{epsfig}

\usepackage{algorithm}
\usepackage{algpseudocode}

\DeclareMathOperator*{\argmax}{argmax}

\begin{document}

\title{Robotic Wireless Networks in a Narrow Alley: \\ A Game Theoretic Approach}

\author{\IEEEauthorblockN{Taehyoung Shim and Seong-Lyun Kim}\\
\IEEEauthorblockA{Department of Electrical and Electronic Engineering, Yonsei University \\
50 Yonsei-Ro, Seodaemun-Gu, Seoul 120-749, Korea  \\
\{teishim, slkim\}@ramo.yonsei.ac.kr}

\thanks{This research was supported by the MKE (The Ministry of Knowledge Economy), Korea, under the ITRC (Information Technology Research Center) support program supervised by the NIPA (National IT Industry Promotion Agency) (NIPA-2012-H0301-12-1001).
}
}

\maketitle

\begin{abstract}
There are many situations where vehicles may compete with each other to maximize their respective utilities.
We consider a narrow alley where two groups, eastbound and westbound, of autonomous vehicles are heading toward each of their destination to minimize their travel distance. However, if the two groups approach the road simultaneously, it will be blocked. 
The main goal of this paper is to investigate how wireless communications among the vehicles can lead the solution near to Pareto optimum. In addition, we implemented such a vehicular test-bed, composed of networked robots that have an infrared sensor, a DC motor, and a wireless communication module: ZigBee (IEEE 802.15.4).

\end{abstract}

\begin{IEEEkeywords}
Wireless Ad Hoc Communications, Vehicle-to-Vehicle (V2V) Communications, Robotic Networks, Non-cooperative Game Theory, Test-bed
\end{IEEEkeywords}


\section{Introduction}\label{section:introduction}


These days, vehicular communications emerged out of industrial technologies and opened a global market to solve a road traffic problem
\cite{Saad:2011}. 
Most of existing work in vehicle-to-vehicle (V2V) communications and vehicle-to-infrastructure (V2I) communications can enhance a safety precaution and an efficiency road transport system such as preventing accident, and reducing traffic congestion via the wireless communications. 
Particularly, in autonomous vehicular networks the wireless communications can be used as sharing information to collaborate among the vehicles \cite{Iqbal:2009}. 
In \cite{slkim:2010}, the networked robots based on ZigBee (IEEE 802.15.4) are able to communicate with other colleague robots by using the wireless multihop communications to find an exit path. In order to solve a minimum-time surveillance problem, the autonomous vehicles also collaborate with others via the wireless communications \cite{Anisi:2010}. 
These autonomous vehicles have to share infrastructure like the road with others and they have limited resources, such as a battery life, a time constraint and so on. For these reasons, the autonomous vehicles compete with each other to maximize their utility in a distributed system. 

In this paper, we suggest a vehicular game theoretic model for a narrow alley. The width of road is limited to only one direction like a one-way street, but it is faster than a detour. So the selfish autonomous vehicles use this fast but busy way. If two groups of autonomous vehicles in the narrow alley are heading toward each of their destinations simultaneously, then the road will be blocked by themselves. 
The utility of each vehicle is maximized when its escaping time from the road is minimized. With a distributed and parallel manner, the vehicles are able to achieve near to Pareto optimum using the wireless ad hoc communications among the vehicles.

The remainder of this paper is organized as follows. First, we introduce our problem and find a solution by using a central authority.
Second, we design our distributed non-cooperative and one-shot game theoretic algorithm from the two-vehicle to the multiple-vehicle problem.
Third, we describe how we handled the game theoretic problem on our ZigBee based networked robot test-bed. Then we show our simulation results. Finally, we conclude the paper.\\

\section{Problem Formulation}\label{section:problem formulation}

\begin{figure}[!t]\centering
  {\includegraphics[scale=0.33]{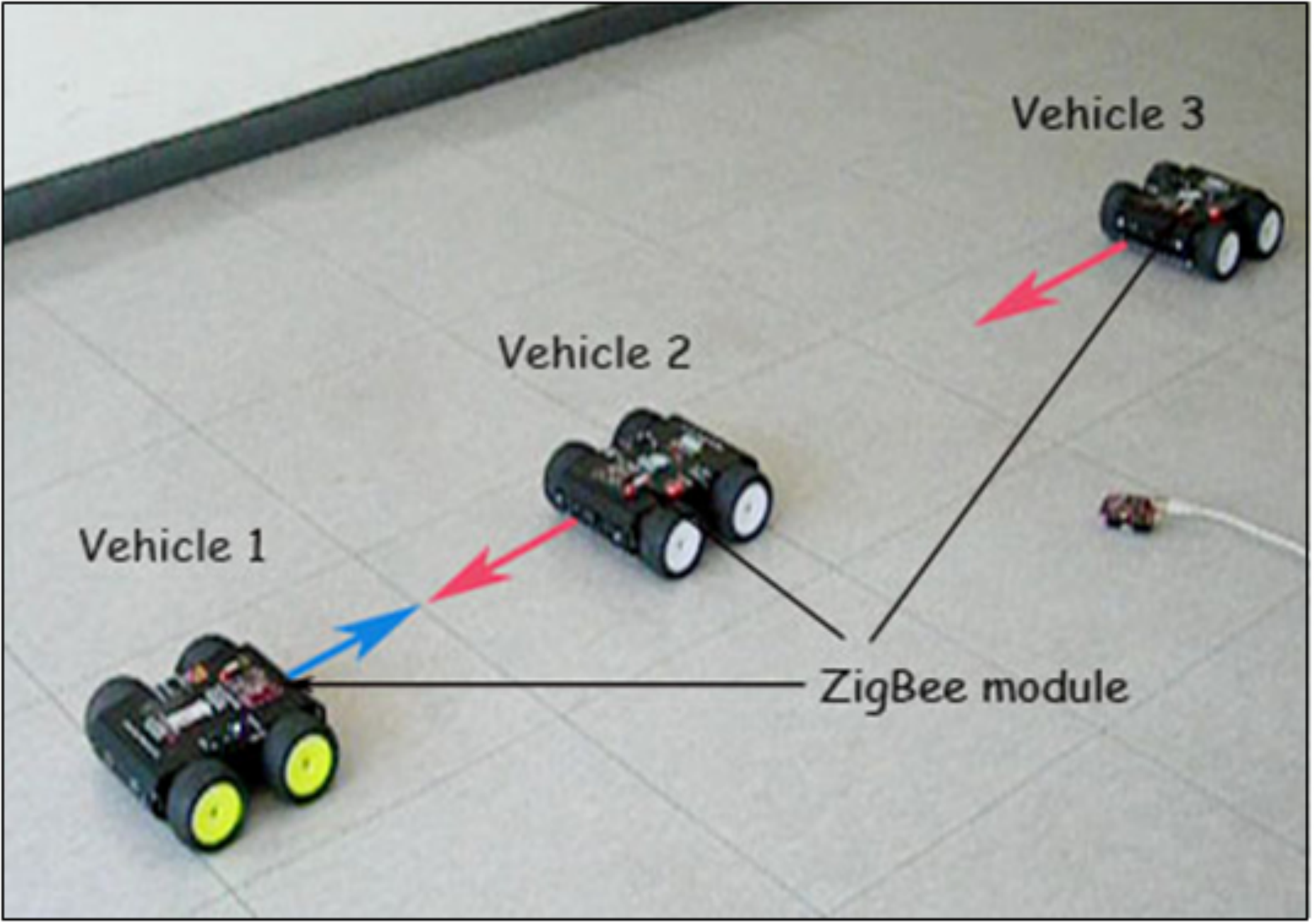}}
  \caption{The test-bed of three vehicles in the narrow alley}\label{fig_implementation}
\end{figure} 

We assume that there are $I$ vehicles in the length of the narrow alley $L$. The action set is $A =  \left\{ +1, 0, -1 \right\}$. In this road the vehicle $i$ chooses the action, $a_i \in A$, each time slot.
We consider minimizing the maximum elapsed time problem in the road where two groups, eastbound and westbound, of autonomous vehicles are heading toward each of their destinations. The problem is described as follows:

\begin{equation}\label{eq_ETM_problem}
\mathop {\min }\limits_{\left\{ {a_1, a_2, \cdot  \cdot  \cdot ,a_{|I|} }\right\}} 
\max \left\{ 
{f_1 \left( {a_1 } \right)}, 	{f_2 \left( {a_2 } \right)}, 	\cdot  \cdot  \cdot, {f_{|I|} \left( {a_{|I|} }
\right)} 
\right\}
,\end{equation}
where $f_i \left( a_i \right)$ denotes the expected elapsed time of vehicle $i$ when it chooses the action $a_i$.

If we assume that all of vehicles are controlled by a central authority, then the problem of (\ref{eq_ETM_problem}) could be solved as follows:
\begin{equation}\label{eq_EET_solution}
\begin{array}{l}
 f_i \left( {a_i } \right) = t_{e,i}  + \left( {L - x_i } \right) + \left( {\frac{{1 - a_i }}{2}} \right)\left( {L - x_{j^*}  + x_i } \right)
,\end{array}
\end{equation}
where $t_{e,i}$ and $x_i$ are the elapsed time of vehicle $i$ and the passed distance of vehicle $i$, which is the summation of the actions from the initial to the current time, respectively. $x_{j^*}$ is the passed distance of the last vehicle among the vehicle $i$'s opponents. 
The central authority finds the maximum passed distance, $x_{\max}$. If the group contains the vehicle having $x_{\max}$, then the member of this group is defined as the high priority group. On the contrary, the other group has the low priority. 
Depending on the priority, the central authority controls all of the vehicles. If the vehicle $i$ has the high priority, then the central authority gives it the forwarding action, $a_i=+1$ and calculates the expected elapsed time of vehicle $i$ using (\ref{eq_EET_solution}). Likewise, if the vehicle $i$ is the member of the low priority, then the authority orders it to have the backwarding action, $a_i=-1$. The last escaping vehicle of the group having the low priority is the solution of the problem of (\ref{eq_ETM_problem}).



Each the utility of vehicle is maximized when its escaping time from the narrow alley is minimized possibly.
In the next section to find the solution of (\ref{eq_ETM_problem}) in a distributed manner, we take a game theoretic approach that leads to a solution near to Pareto optimum.



\section{Game Theoretic Approach}\label{section:mobility control}

\begin{figure}[t!]\centering
  {\includegraphics[scale=0.56]{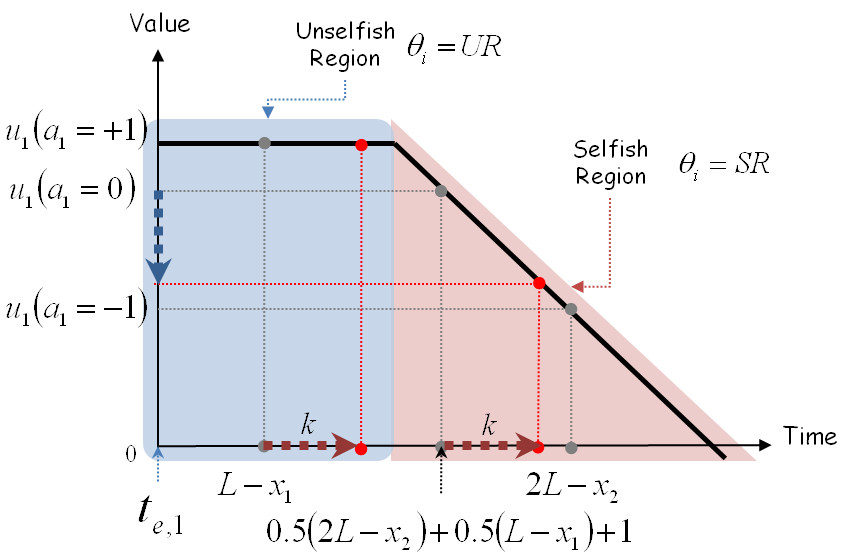}}
  \caption{ The utility function is composed of the block shape and the triangle shape. The type of vehicle $i$ is high in the unselfish region, but it is low in the selfish region.}
  \label{fig_utility_function}
\end{figure}

Each of vehicles has a utility function which is composed of its type and the expected elapsed time, $u_i (\theta_i, f_i)$. In Figure \ref{fig_utility_function}, the shape of the utility function is described as a block and a triangle shape. The former is the unselfish region where they would make way for the others; the latter is the selfish region where they choose the action to maximize their payoff selfishly. 
The type of vehicle $i$ is high, $\theta_i = UR$, in the unselfish region. On the contray, it is low, $\theta_i = SR$, in the selfish region. 
In this section we introduce a non-cooperative and one-shot game from the two-vehicle to the multi-player game, and design the game theoretic algorithm for the autonomous vehicles.

\subsection{Two-Vehicle Game}
\begin{figure}[t!]\centering
 {\includegraphics[scale=0.8]{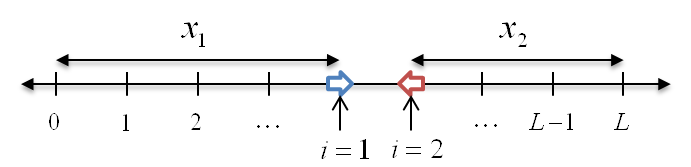}}
  \caption{The narrow alley networks where two groups of autonomous vehicles are heading toward opposite directions. The eastbound vehicle $1$ has the passed distance, $x_1$, from starting point, $0$. The westbound vehicle 2 has the passed distance, $x_2$, from starting point, $L$. }
  \label{fig_narrowalleymodel}
\end{figure}

We assume that there are two vehicles, $I = \{1,2 \}$, which are shown in Figure \ref{fig_narrowalleymodel}. The eastbound vehicle $1$ has the passed distance, $x_1$, from starting point, $0$ and the westbound vehicle 2 has the passed distance, $x_2$, from starting point, $L$. 

The vehicle $1$'s best respons to pure strategy profile in each time slot is given by
\begin{equation}\label{eq_best_response}
 a_1^* = \argmax_{a_1 \in A } E \left[ u_1 \left(\theta_1, f_1 \left(a_1 \right) \right) \right],
\end{equation}
where $a_1^*$ denotes the best action of the vehicle $1$. If there are no pure strategy in this game, then each player chooses a set of mixed strategies. The set of vehicle i's mixed strategy is defined by the set of probability distributions over possible actions, $p(A_1) = \{ p\left( a_1 =+1 \right),  p\left( a_1 =0 \right),  p\left( a_1 =-1 \right) \}$.



In the two-vehicle game, the expected elapsed time of vehicle $1$, $f_1$, depends on its decision. 
If the vehicle $1$ chooses the forwarding action, its expected elapsed time, $f_1$ depends only on the rest of the way, $L-x_1$. If the vehicle $1$ selects the backwarding action, then $f_1$ is similarly calculated by the rest time of competitor, $2L - x_2$. The expected elapsed time for the waiting action is composed of the turning and the thinking time, $0.5\left(L - x_1\right)+0.5\left(2L - x_2\right)+1$, because the vehicle $1$ does nothing during the time slot. 


In Figure \ref{fig_narrowalleymodel}, if the vehicle $1$ and $2$ have the forwarding action, then the collision occurred between them. The time cost between the two vehicles $t_{col}$ is defined as $t_{col} =   {k\,},  k \in \left[ {0,\infty } \right)$ where $k$ is time cost rate that is added to their elased time . 
In Figure \ref{fig_utility_function}, if the vehicle $1$ chooses the forwarding strategy, then its payoff is in the unselfish region. However, if vehicle $1$ chooses the others, the backwarding or the waiting strategy, then its payoff is in the selfish region.

\subsection{Multi-Vehicle Game}
We assume that there are vehicles more than two-vehicle in the narrow alley. We assume that there are vehicle $i+1$ and $i-1$, in front and back of the vehicle $i$, respectively. The vehicle $i$ in the middle of them has to consider the type of neighbors. Its best action is defined as follow:
\begin{equation}\label{eq_best_response2}
 a_i^* = \argmax_{a_i \in A } \sum_{ \theta_{~-i} } \sum_{ a_{~-i}} 
p\left( \theta_{~-i}, a_{~-i} \right) u_i \left(\theta_i, f_i \right), 
\end{equation}
where ~-$i$ denotes the vehicles in this road except the vehicle $i$. If we could find the utility function of multi-vehicle payoff matrix, then we can calculate the best-response strategy which is the solution near to Pareto optimum.

\subsection{Game Theoretic Algorithm}
In the autonomous vehicular networks, there exist a sensing range of each other between the vehicle $1$ and $2$, i.e. $\left| {s_1  - s_{2} } \right| < D\left( {s_i } \right)$, where $s_i$ and $D(s_i)$ denote the state of vehicle $i$ and the sensing range of vehicle $i$ on the state $s_1$ respectively.
Two vehicles can detect each other with the distance measuring sensor, for instance the infrared or the ultrasonic sensor.


If each vehicle cannot share the information, for example the types, the elapsed time and so on, then the vehicle has to predict the possible strategy using only its own. 
The vehicles in this road choose its action to maximize its utility without the clues of neighbors. Assuming that each vehicle, commonly, chooses a conventional vehicle movement by the non-cooperative game, we will expect the action of another vehicle stochastically.
According to the outcomes of each vehicle, it decides its action which is to be minimize the expected elapsed time and maximize its utility simultaneously. 


\begin{figure}[!t]\centering
  {\includegraphics[width=2.8in]{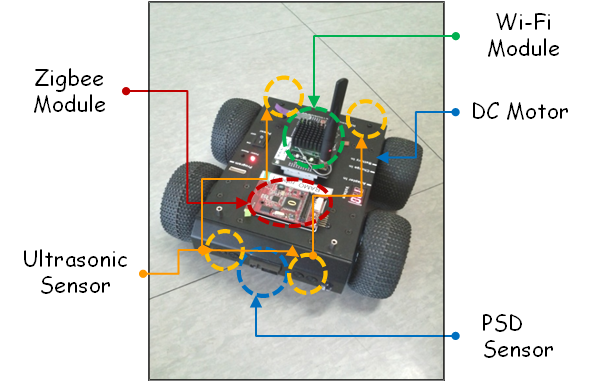}}
  \caption{The networked robot with the wireless communication module}\label{fig_networked_robot}
\end{figure} 

If they can share information with the others, such as the type of others, then the uncertainty of the action of others can be reduced in th this model via wireless communications. Therefore, we investigate how the communications among the vehicles can lead the solution near to Pareto optimum where no one can be made better off by making someone worse off. In next section, we implemented such a vehicle test-bed to show the gain of wireless communications, which is based on the game theoretic algorithm. The algorithm is distributed and selfish. It is composed of the forwarding part and the game theoretic strategy part. If the vehicle doesnot sense the other vehicles, then it has the forwarding action. If the vehicle can detect the others, then they share the type information and choice the action which is maximizing their utility function.

\begin{algorithm}
\caption{Game Algorithm With Communications}\label{alg_gametheory}
\begin{algorithmic}[1]
\Procedure{Non-cooperative Game}{$ S, \Theta $}
  \If {$ |s_i - s_{-i}| \leq D\left( s_i \right) $} \Comment{Detecting an object}
     \State $a_i^* = \argmax_{a_i \in A} p\left( a_{-i} \right) u_i \left(\theta_i, a_i\right)$
   \Else
     \State $a_i \gets +1$  \Comment{ Forward action}
 \EndIf
\EndProcedure
\end{algorithmic}
\end{algorithm}



\begin{table}[t!]
\renewcommand{\arraystretch}{1.0}
\centering
\begin{tabular}{   c  c    }
\hline
  \bf Parameter    &  \bf Value \\
\hline
   Frequency band   &  2.4 GHz \\
   Max signal rate   &  250 Kb\//s \\
  TX range  &   10 -- 100  m  \\
   TX power    &   {-25} -- 0 dBm \\
   Channel bandwidth    &   2 MHz  \\
   Cell size    & 450 $\times$ 450 mm \\
   Motor torque & 2.0 Kg\//cm \\
  Motor speed & 102 rpm \\
   Distance measuring range &  10 -- 80 cm \\
  Flash Memory  & 64 MByte\\
\hline
\end{tabular}
\caption{Experimental Environments} \label{tb_experiment}
\end{table}

\section{Implementation}\label{section:implementation}
\subsection{Robot Structure}
We have implemented our non-coperative game theoretic algorithm on networked robots, which conforms to ZigBee (IEEE 802.15.4) physical layer specifications. Each robot is equipped with an 8-bit AVR ATmega128 and a CC2420 chip as a microcontroller and a radio transceiver, respectively as in Figure \ref{fig_networked_robot}. The radio transceiver has the PCB antenna. We also use the four powerful DC motor and the distance measuring sensor which has the distance range up to 80cm. We define the cell size, 450mm, which is less than the the measuring maximum distance. All of robots have the same utility function which has the unselfish and selfish region.  In order to synchronize the current time in this network, we use the assistance node who broadcast the time slot message. We also define the length of time slot which has 3s for movement and 1s for communication. As in Figure \ref{fig_implementation}, we implemented our algorithm on the three-vehicle networks.

\subsection{Communication System}
ZigBee (IEEE 802.15.4) module is easy to construct networks because of the advantages of ZigBee, for example, a self-organization, a multi-hop communications and a long battery lifetime which has a maximum rate $250 Kb\//s$ that is sufficient to share all of information in the networks. In wireless communications aspect, ZigBee MAC protocol has CSMA/CA, random backoff and packet transmit type; DATA or ACK \cite{JSLee:2007}. In our wireless communication test-bed, we use a broadcasting message which is sent once and then repeated by the neighborings and check the source address and the sequence number of the packet. Our experimental environments are shown in Table \ref{tb_experiment}.\footnote[1]{Moving pictures of the experiments can be downloaded from  $http://hertz.yonsei.ac.kr/Narrow\_Alley.wmv$ .} From this implementation, we investigate how wireless communications among the vehicles can lead the solution near to Pareto optimum


\section{Simulation Results}\label{section:simulation result}

\begin{figure}[!t]\centering
  {\includegraphics[width=3.3in]{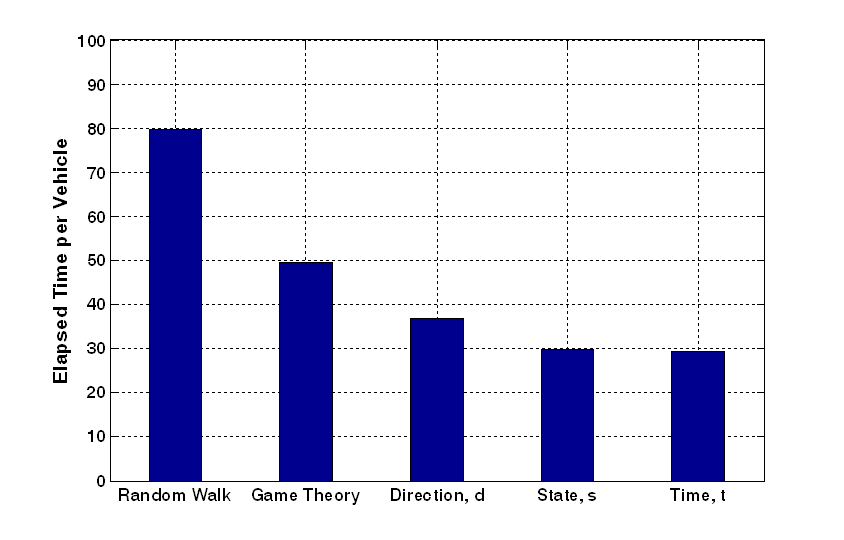}}
  \caption{The elapsed time per vehicle in the narrow alley }\label{fig_ET}
\end{figure}

In this section, we present that the gain of wireless communications in Figure \ref{fig_ET}. In this simulation, we set the length of the narrow alley $L=20$, and the collision cost rate $k=10$. Each of vehicles can share the information via ad hoc networking, reduce the average elapsed time per vehicle by 37.8\% and 54.0\%, respectively, comparing to a conventional vehicle movement (random) algorithm. By adding the current state information to the transmitted packets, furthermore, the average escape time can be reduced by 62.4\%.

\begin{figure}[!t]\centering
  {\includegraphics[width=3.3in]{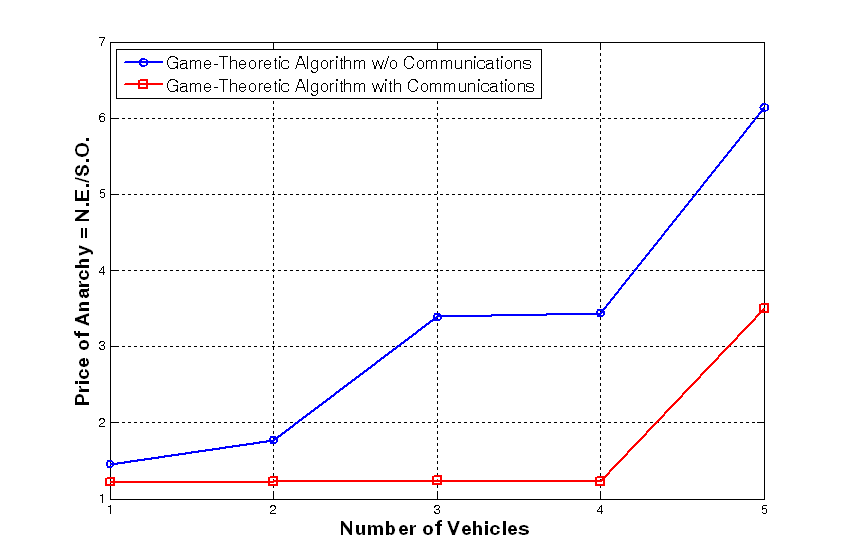}}
  \caption{The Price of Anarchy of  game theoretic algorithm without communications versus with communications compared with the social optimum}\label{fig_PoA}
\end{figure}

In Figure \ref{fig_PoA}, we extend to the multiple vehicular networks in this road. If we use the game-theoretic algorithm without the wireless communications to our networks, then the Nash equilibrium is almost up to six times than the social optimum. On the contrary, if we use the wireless communication among the vehicles in this road, then we can reduce the congestion price that closed to Pareto optimum up to four vehicles.\\

\section{Conclusions and Remarks}\label{section:conclusion}
In this paper, we suggest a non-cooperative game theoretic model that is to be maximized unselfishly. 
We investigate how the wireless communications among the vehicles, can lead the solution to the nearly Pareto optimum. 
we presented simulation that show the gain of wireless communications. 
Our game theoretic algorithm can reduce the average escape time per vehicle by 37.8\%. 
And the information sharing via ad hoc networks can reduce the average escape time per vehicle by 54.0\%. 
By adding the current state information to the transmitted packets, furthermore, the average escape time can be reduced by 62.4\%. 
In the multiple vehicular networks, we can reduce the congestion price that closed to Pareto optimum up to 4 vehicles using the wireless communication among the vehicles. 
Based on game theoretic approach we implemented such the vehicular test-bed, composed of networked robots. \\

\end{document}